\documentclass[letterpaper, 10 pt, conference]{ieeeconf}  

\IEEEoverridecommandlockouts                              
\usepackage{times}
\usepackage{latexsym}
\usepackage[T1]{fontenc}
\usepackage[utf8]{inputenc}
\usepackage{microtype}
\usepackage{inconsolata}
\usepackage{graphicx}
\usepackage{booktabs}
\usepackage{multicol}

\usepackage{listings}
\usepackage{color}
\title{\LARGE \bf
FurNav: Development and Preliminary Study of a Robot Direction Giver 
}

\author{Bruce W. Wilson$^{1*\dag}$, Yann Schlosser$^{1}$, Rayane Tarkany$^{1}$, Meriam Moujahid$^{1}$, \\ Birthe Nesset$^{1}$, Tanvi Dinkar$^{1\ddag}$, and Verena Rieser$^{1,2\ddag}$
\thanks{$^{\dag}${ PhD funding is provided by Engineering and Physical Sciences Research Council (EPSRC) Centre for Doctoral Training in Robotics and Autonomous Systems.}}%
\thanks{$^{\ddag}${Tanvi Dinkar and Verena Rieser were supported by the EPSRC project `Gender Bias in Conversational AI' (EP/T023767/1), `Equally Safe Online' (EP/W025493/1), and `AISEC: AI Secure and Explainable by Construction' (EP/T026952/1), Verena Rieser was also supported by a Leverhulme Trust Senior Research Fellowship (SRF/R1/201100).}}
\thanks{$^{*}${Corresponding Author: \tt\small bww1@hw.ac.uk}}%
\thanks{$^{1}$School of Mathematical and Computer Sciences,
        Heriot-Watt University, UK}%
\thanks{$^{2}$Now at Google DeepMind}%
}

\begin{document}

\maketitle
\thispagestyle{empty}
\pagestyle{empty}

\begin{abstract}


When giving directions to a lost-looking tourist, would you first reference the street-names, cardinal directions, landmarks, or simply tell them to walk five hundred metres in one direction then turn left? 
Depending on the circumstances, one could reasonably make use of any of these direction giving styles. However, research on direction giving with a robot does not often look at how these different direction styles impact perceptions of the robots intelligence, nor does it take into account how users prior dispositions may impact ratings.
In this work, we look at generating natural language for two navigation styles using a created system for a Furhat robot, before measuring perceived intelligence and animacy alongside users prior dispositions to robots in a small preliminary study (\textit{N}=7).
Our results confirm findings by previous work that prior negative attitudes towards robots correlates negatively with propensity to trust robots, and also suggests avenues for future research. 
For example, more data is needed to explore the link between perceived intelligence and direction style.
We end by discussing our plan to run a larger scale experiment, and how to improve our existing study design.
\end{abstract}

\section{INTRODUCTION}

We take common ground for granted in many human-to-human interactions.
When for example walking up to an airport security agent and asking ``where is gate 10?'', both interlocutors understand the context of the situation, knowing where they are, the appropriate topics of conversation, who has authority, and approximately who has knowledge of what \cite{kiesler_fostering_2005}.
More concretely, this \emph{common ground} relies on a level of shared knowledge between involved parties \cite{clark_context_2006}, and, to form this, comparable mental models of one another must be created.

As such, an interactive robot should be equipped with common ground capabilities to achieve effective communication \cite{kiesler_fostering_2005}. This is particularly applicable to a \emph{situated task}, where expressions used in a dialogue have an interdependence on the immediate environment.
With this statement includes interactive direction giving robots, where just like in the ``where is gate 10?'' example above, a level of common ground would greatly improve communication. However, this common ground could be achieved in multiple ways. Landmarks are one potential avenue, which have been suggested to improve the navigational efficiency and reliability of route instructions \cite{denis_spatial_1999}.  

\begin{figure}[t]
\centerline{\includegraphics[scale=0.15]{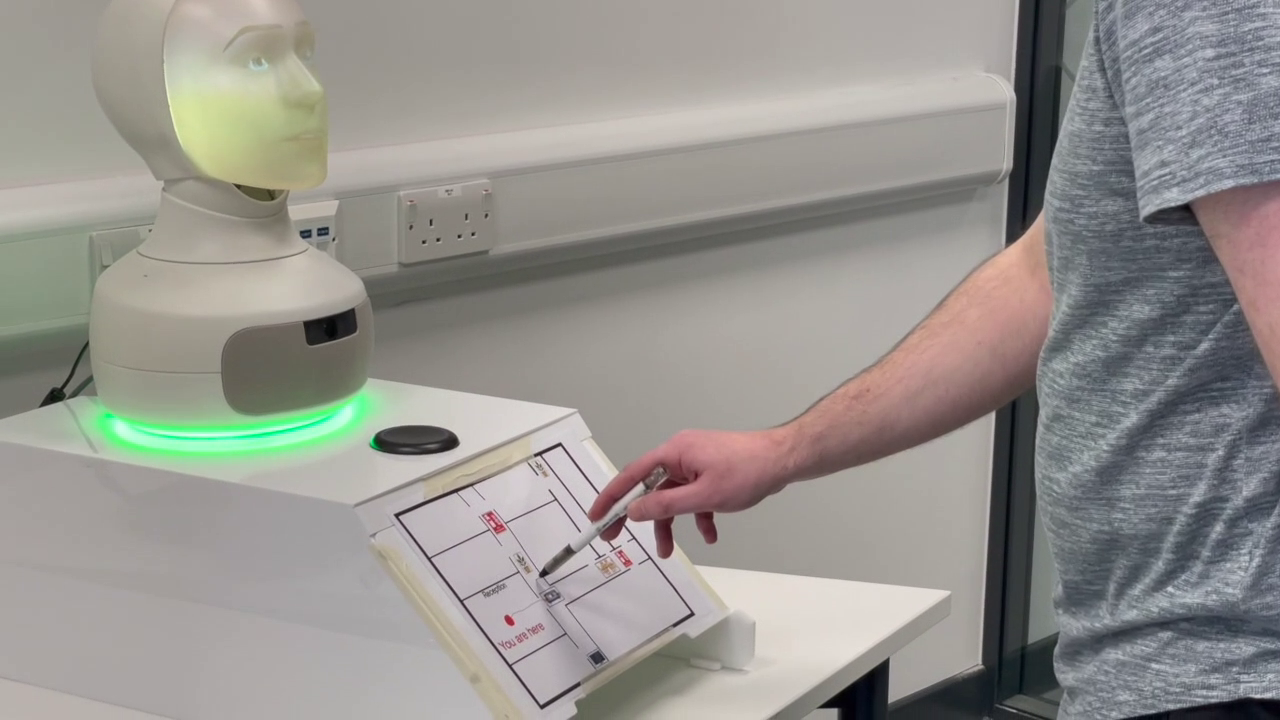}}
\caption{The Furhat direction giver setup with a user interacting with the preliminary study system. The map shown is visible in Figure \ref{fig:map}.}
\label{fig:userdrawing}
\end{figure}

In this paper, we present a methodology and preliminary experiment with our robot direction giver in a lab setting, shown in Figure \ref{fig:userdrawing}. A Furhat robot\footnote{https://furhatrobotics.com/} is set up to provide navigation instructions in one of two conditions: landmark or skeletal instruction based directions. Participants will navigate around a map based on these instructions, drawing their path with a pen.
Based on this setup, we focused on whether the use of landmark-based directions, and in turn, 
an assumed level of common ground with the user, impacts users perceived intelligence and animacy rating of the robot using the Godspeed questionnaire sub-scales \cite{bartneck_measurement_2009}. 
We also factor in users' prior attitudes towards robots, specifically their propensity to trust robots \cite{jessup_measurement_2019}, and their negative attitudes towards robots \cite{nomura_measurement_2006}. 
Thus in this preliminary study, we aim to answer the following research questions:
\begin{itemize}
    \item RQ1: Does the assumption of common ground for a navigation-based task influence perceived intelligence and animacy rating of the robot?
    \item RQ2: What factors play a role in the perceived intelligence and animacy of the robot?
\end{itemize}

\section{RELATED WORK}

While navigating, a person may use various spatial, cognitive, and behavioral abilities to be able to find their way along a route \cite{raubal_enriching_2002}. 
This route is usually split into several segments which can individually be verbalised \cite{couclelis_verbal_1996}, either referring to particular actions such as ``turn'', ``walk'', or environmental descriptors such as a ``red car'', often accompanied with a skeletal direction ``to your left'', helping to aid identification of where an action should be carried out \cite{allen_knowledge_1997}. Action or direction order should be reflective of the linear order that the route is to be traversed \cite{allen_principles_2000}. Concretely, each instruction step can be split into two main components which contribute to distinct functions in the discourse and must be viewed separately: a procedural action that a navigator should perform, e.g., ``turn right''; and a description of where in the environment the action should be executed ``to the right of the church'' \cite{tom_language_2004}.

In particular, several studies have pointed to the fact that landmarks play a crucial role in communicating route directions \cite{michon_when_2001, caduff_landmark_2005}. For example, it is much easier for a navigator to find their way if they can rely on a description of the route based on well-recognisable objects in their environment (rather than relying on street names and metric directions alone) \cite{tom_referring_2003}. Clarity of specific route instructions is also improved with landmarks, improving navigation efficiency and reliability \cite{denis_spatial_1999}.
Finally, they can be used identify critical decision making points along the route, for example, where a turning action has to be taken \cite{michon_when_2001}.

\begin{figure}[b]
    \centering
    \includegraphics[scale = 0.4]{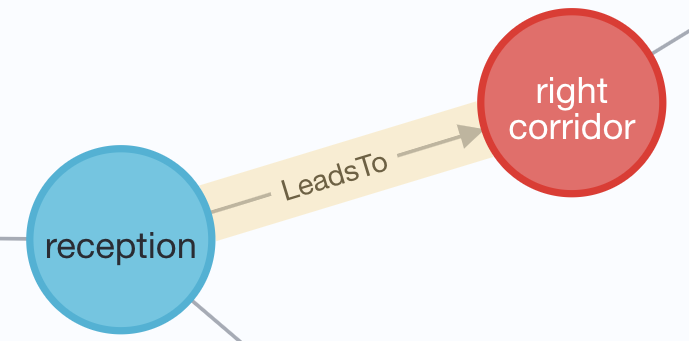}
    \includegraphics[scale = 0.4]{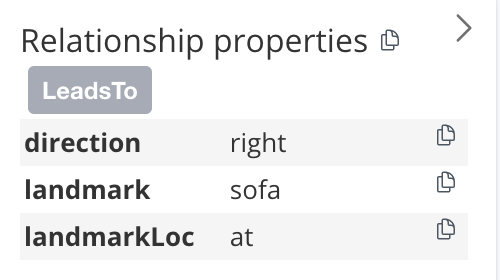}
    \caption{Two nodes representing rooms, a relationship between them indicating the traversable route, and the data stored in the relationship's properties (e.g. landmark at that decision point).}
    \label{fig:graphandproperties}
\end{figure}

On the other hand, skeletal based navigation, described by \cite{denis_description_1997}, involves abstractions of navigation instructions, reflecting the essence of each route distilled from actual route descriptions. This, for example, may involve the route steps of simply ``go forward, turn left, go forward, then turn right,'' which still contain the essential of the navigational procedure, however do not contain any extra embellishment.

In human-to-human interactions, the way in which spacial knowledge is communicated through routes and instructions has been extensively studied, (for example, see \cite{allen_principles_2000}).
\cite{cercas-curry-etal-2015-generating} evaluates where landmarks may be helpful in a virtual interactive environment, taking into account where a landmark may not be visible on a route. They found that their heuristic-based approach, taking into account visibility, outperforms two-corpus based systems in terms of naturalness and task completion, however their results were not significant.
Researchers have also looked at this issue from a human-robot perspective. \cite{heikkila_should_2019} looks at a robot situated in a shopping center, with one of its abilities being to give humans route guidance. After conducting a four-phased qualitative study, they found nine design implications, one of which noted that salient landmarks and those located in the crossings of aisles are helpful, but one must moderate their use.  
Finally, \cite{bohus_directions_2014} looks at providing natural language directions in an in-the-wild experiment, finding that including landmarks may be useful navigational way-points for longer routes.

        
\section{Study Methods}

\subsection{Direction Generation}

\begin{figure}[h]
\centerline{\includegraphics[scale=0.11]{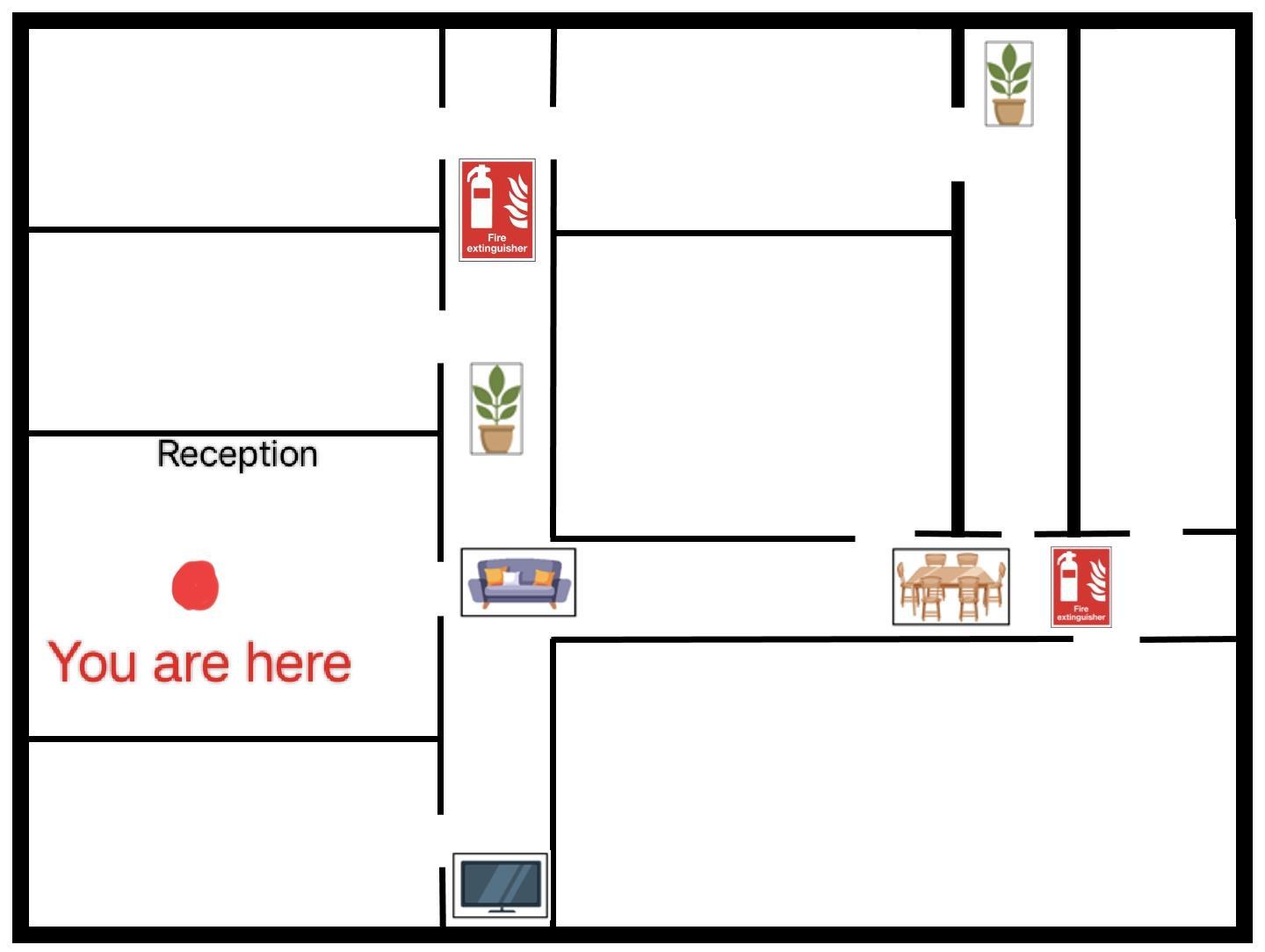}}
\caption{Our created map showing the starting location (reception), alongside visible landmarks, corridors, and rooms. The generated text will guide participants around the map as if they were actually walking the direction.}
\label{fig:map}
\end{figure}

In order to generate either landmark or skeletal route instructions for navigation, we required a map and knowledge base. 
Firstly, the map, shown in Figure \ref{fig:map} was created with a consistent starting location, and numbered rooms along numerous corridors, with rooms representing destinations. Landmarks were placed at each decision making point. Rooms are only labelled in the knowledge base, and not on the map given to participants.

This map was then used to create a knowledge base in the form of a neo4j\footnote{https://neo4j.com/} graph database. In this database, nodes represent rooms and corridors, and relationships between nodes contain properties relevant to either skeletal or landmark based directions at decision making points. These properties are either turning directions, used in both types of route instructions, or the landmark placed at the decision making point, used only in landmark type route instructions. An example of these nodes and properties is shown in Figure \ref{fig:graphandproperties}.

This knowledge base is then queried using Cypher\footnote{https://neo4j.com/docs/getting-started/cypher-intro/}, a query language created for neo4j graph databases. The shortest path metadata is then extracted, following from the starting location to the destination, moving along corridors and turns as appropriate. 
From this metadata, natural language is generated using a template-based approach, where each section of the route is constructed by randomly selecting a template and filling it with the appropriate landmark or skeletal based instruction. 

From this, if you were to ask for room four (the room below the reception), an example of the generated text would be:
\begin{itemize}
    \item Landmark - Turn right in the corridor at the sofa. Follow the corridor and turn right at the TV.
    \item Skeletal - Go right in the corridor. Follow the hallway and turn right.
\end{itemize}

\begin{table*}[h]
	\centering
    \hspace{1pt}
	\caption{Descriptive statistics for all numeric measures (\textit{N}=7). Cronbach's $\alpha$ is used to measure internal consistency of the sub-scale items, where a higher value denotes higher measured internal consistency ($\uparrow$).}
	\label{tab:descriptiveStatistics}
	{
		\begin{tabular}{lrrrrrr}
			\toprule
			 & Median & Mean & Std. Deviation & Minimum & Maximum & Cronbach's $\alpha$ ($\uparrow$) \\
			\cmidrule[0.4pt]{1-7}
			NARS Score & $2.214$ & $2.214$ & $0.389$ & $1.429$ & $2.643$ & - \\
			PTT Score & $3.333$ & $3.595$ & $0.615$ & $3.000$ & $4.833$ & - \\
			Skeletal Godspeed Animacy Score & $2.400$ & $2.800$ & $0.849$ & $2.000$ & $4.200$ & $0.85$  \\
			Skeletal Godspeed Intelligence Score & $3.400$ & $3.400$ & $0.833$ & $2.400$ & $4.400$ & $0.80$  \\
			Landmark Godspeed Animacy Score & $3.800$ & $3.686$ & $0.363$ & $3.200$ & $4.200$ & $0.13$  \\
			Landmark Godspeed Intelligence Score & $3.800$ & $3.743$ & $0.550$ & $3.000$ & $4.600$ & $0.75$  \\
			Skeletal Success Rate & $2.000$ & $1.571$ & $1.272$ & $0.000$ & $3.000$ & - \\
			Landmark Success Rate & $3.000$ & $2.286$ & $0.951$ & $1.000$ & $3.000$ & - \\
			\bottomrule
		\end{tabular}
	}
\end{table*}

\subsection{Robot Embedding}

The natural language generation component was then deployed using a Neo4J Kotlin driver onto a Furhat robot. This social robot consists of a head and shoulders with a back-projected face, capable of displaying a range of expressions and gestures, including non-verbal cues.
We additionally made use of the on-board camera and microphone, combining it with the Furhat NLU software.
A rule-based model was created to perform intent recognition and entity extraction, relating to the rooms that users may wish to visit.

\section{User Study}

We performed a preliminary evaluation of our created system in a lab based setting. This involved the user interacting with both conditions on the Furhat robot in a lab environment, collecting both objective and subjective measures.



\subsection{Setup}
The Furhat navigation robot was located in a lab environment without observers, with the participant positioned facing the robot, and a facilitator out of the field of view of the participant. The Furhat robot is placed on a plinth on a table, able to gesture and move its head freely, with the microphone wired to next to the participant for optimal ASR results. This setup is shown in Figure \ref{fig:userdrawing}.
The map is placed on the table in-front of the robot, so that the robot may gesture to it in speech, with a pen available so that the participant can draw on the map as instructed.



\subsection{Experimental Protocol}
A within-subjects study design with a randomised initial condition was used. 
Each condition, skeletal or landmark, contained three tasks: navigating to rooms 5, 3, then 7, sequentially increasing in navigation difficulty.
Participants were instructed to listen and follow along to the navigation steps by drawing a single continuous line from the starting point to their destination.
Rooms were numbered only internally in the knowledge base, and were not numbered on the copy of the map given to participants to draw on.
 
Participants were guided through the experiment with an interactive questionnaire, which first gathered informed consent, before presenting the NARS and PTT questionnaires.
When the participant is ready, the facilitator will begin the experiment, where the Furhat will read out a short introduction explanation explaining the task. Participants will complete the first condition, before being asked to rate the robot on the Godspeed animacy and perceived intelligence sub-scales. After this, the participants will be asked to complete the second condition, followed by the same questionnaires.



\begin{table}[b]
	\centering
	\caption{Pearson's r Correlations for valid metrics (accepted Cronbach's $\alpha$). Accepted p values are $<0.05$ ($\downarrow$). Skl. denotes Skeletal, and Lnd. denotes Landmark}
	\label{tab:pearsonSCorrelations}
	{
		\begin{tabular}{lrrrr}
			\toprule
			 &  &  & Pearson's r & p ($\downarrow$)  \\
			\cmidrule[0.4pt]{1-5}
			NARS Score & - & PTT Score & $-0.912$ & $0.004$  \\
			  NARS Score & - & Skl. Intelligence Score & $0.632$ & $0.128$  \\
			PTT Score & - & Skl. Intelligence Score & $-0.423$ & $0.344$  \\
   		  NARS Score & - & Lnd. Intelligence Score & $-0.489$ & $0.265$  \\
			PTT Score & - & Lnd. Intelligence Score & $0.692$ & $0.085$  \\
			\bottomrule
		\end{tabular}
	}
\end{table}

\subsection{Metrics}

Several objective and subjective measures were collected:

\textbf{Pre-interaction:} Negative Attitude Towards Robots (NARS), Propensity to Trust Technology (PTT).

\textbf{Each Interaction Condition:} Individual task success, perceived intelligence (Godspeed sub-scale), animacy (Godspeed sub-scale).





To test \textbf{RQ1}, we will perform a Wilcoxon signed-rank statistical test with the collected task success, perceived intelligence, and animacy measures.
Similarly, to test \textbf{RQ2}, we perform correlations using the collected NARS and PTT questionnaires against themselves and our collected perceived intelligence scores.

\section{Preliminary Results}

From our study setup, we can analyse the preliminary results, and gather general trends on \textbf{RQ1} and \textbf{RQ2} before our full study.
Table \ref{tab:descriptiveStatistics} gives the descriptive statistics for all the continuous measures mentioned above. 
Cronbach's Alpha was calculated over the Godspeed Questionnaire sub-scales, shown in Table \ref{tab:descriptiveStatistics}. Landmark animacy score returns an $\alpha$ value lower than acceptable in this use case, due to the small sample size. Therefore, animacy comparison was excluded in this preliminary study.


We then compared the mean results from our Godspeed perceived intelligence
and task success with paired sample Wilcoxon signed-rank tests across conditions: (\textbf{Skeletal Godspeed Intelligence Score} Mean = 3.40, \textbf{Landmark Godspeed Intelligence Score} Mean = 3.74, \textit{N}=7, \textit{z}=-0.94, sig two-tailed \textit{p}=0.40),
(\textbf{Skeletal Task Success} Mean = 1.57, \textbf{Landmark Task Success} Mean = 2.29, \textit{N}=7, \textit{z}=-1.83, sig two-tailed \textit{p}=0.09); all of which showed no statistically significant results, and at this stage of the preliminary study, it is not possible to draw any conclusions relating to \textbf{RQ1}.





To calculate our correlation between variables, we then used a Pearson's r test between multiple variables, with the correlation table shown below in Table \ref{tab:pearsonSCorrelations}.
As noted in this table, the only statistically significant result is that the \textbf{NARS Score} is negatively correlated with the \textbf{PTT Score} (Pearson's \textit{r}=-0.912, $\textit{p}\leq0.05$), meaning that participants with a higher negative attitude towards robots have a lower propensity to trust technology, which falls in line with previous work \cite{lim_we_2022}.
\textbf{PTT} correlated with \textbf{Landmark intelligence score} is marginally statistically significant, with a positive correlation showing that a higher propensity to trust technology results in a higher average rating of the landmark navigation condition's intelligence score. Overall, \textbf{RQ2} cannot be conclusively answered without further work.

\section{Discussion and future work}

In this paper, we created a system for a direction giving Furhat robot, capable of supplying these directions in two styles, landmark or skeletal. From this, we ran a small scale preliminary study on this system, resulting in statistically insignificant results. However, we plan to run a larger scale experiment using the same system created here with improvements. Based on a power analysis performed using G*Power \cite{Faul2009} with an estimated effect size of $0.42$, to achieve a power (1 - $\beta$) of $80\%$, the required total sample size would be at least \textit{N}=50 participants for an actual power of $81\%$.
Additionally, we would like to switch from measuring perceived intelligence to measuring perceived social intelligence to closer link to existing work on common ground, using for example the PSI Scales \cite{barchard_measuring_2020}.
Moreover, we would like to collect more objective measures, such as task time, clarification requests, and specifics on wrong destinations. We also intend to look at expanding the knowledge base and map to cover the National Robotarium at Heriot-Watt University, where eventually an in-the-wild study could be ran with the Furhat direction giver acting as a robot receptionist. 





\section*{ACKNOWLEDGMENT}
The authors would like to thank Jose Berlin Durai Yoseppu for his work on the development of the Furhat NLU system, alongside the group members for the F21CA class.

\bibliographystyle{IEEEtran}
\bibliography{IEEEabrv,IEEEexample,references}

\end{document}